\documentclass[11pt,a4paper]{article}
\usepackage[hyperref]{acl2021}
\usepackage{times}
\usepackage{latexsym}

\usepackage{microtype}

\aclfinalcopy 
\def\aclpaperid{Submission ID: 1359} 

\usepackage{times}  
\usepackage{helvet} 
\usepackage{courier}  
\usepackage{graphicx} 
\usepackage{natbib}  
\usepackage{caption} 

\usepackage{amsmath}
\usepackage{multirow}
\usepackage{makecell}
\usepackage{arydshln}
\newcommand{\para}[1]{\vspace{0.05in}\noindent\textbf{#1 }}
\usepackage{float}

\usepackage{xcolor}

\title{Learning from Miscellaneous Other-Class Words \\ for Few-shot Named Entity Recognition}
\author{Meihan Tong\textsuperscript{\rm 1}, Shuai Wang\textsuperscript{\rm 2}, Bin Xu\textsuperscript{\rm 1}\thanks{Corresponding author.},  Yixin Cao\textsuperscript{\rm 3}, Minghui Liu{\rm 1}, Lei Hou\textsuperscript{\rm 1}, Juanzi Li\textsuperscript{\rm 1} \\
\textsuperscript{\rm 1}Knowledge Engineering Laboratory, Tsinghua University, Beijing, China\\
\textsuperscript{\rm 2}SLP Group, AI Technology Department, JOYY Inc, China \\
\textsuperscript{\rm 3}S-Lab Nanyang Technological University, Singapore\\
\texttt{tongmeihan@gmail.com, wangshuai1@yy.com}\\ \texttt{ xubin@tsinghua.edu.cn, caoyixin2011@gmail.com}\\
\texttt{liu-mh16@mails.tsinghua.edu.cn,greener2009@gmail.com}\\ \texttt{lijuanzi@tsinghua.edu.cn}}

\date{}

\begin{document}
\maketitle
\begin{abstract}
Few-shot Named Entity Recognition (NER) exploits only a handful of annotations to identify and classify named entity mentions. Prototypical network shows superior performance on few-shot NER. However, existing prototypical methods fail to differentiate rich semantics in other-class words, which will aggravate overfitting under few shot scenario. To address the issue, we propose a novel model, \textbf{M}ining \textbf{U}ndefined \textbf{C}lasses from \textbf{O}ther-class (MUCO), that can automatically induce different undefined classes from the other class to improve few-shot NER. With these extra-labeled undefined classes, our method will improve the discriminative ability of NER classifier and enhance the understanding of predefined classes with stand-by semantic knowledge. Experimental results demonstrate that our model outperforms five state-of-the-art models in both 1-shot and 5-shots settings on four NER benchmarks. We will release the code upon acceptance. The source code is released on \url{https://github.com/shuaiwa16/OtherClassNER.git}.
\end{abstract}
 

\section{Introduction}
Named Entity Recognition (NER) seeks to locate and classify named entities from sentences into predefined classes \cite{yadav2019survey}. Humans can immediately recognize new entity types given just one or a few examples\cite{lake2015human}. Although neural NER networks have achieved superior performance when provided large-scale of training examples \cite{li2019dice}, it remains a non-trivial task to learn from limited new samples, also known as few-shot NER \cite{fritzler2019few}.

Traditional NER models, such as LSTM+CRF \cite{lample-etal-2016-neural}, fail in few-shot settings. They calculate the transition probability matrix based on statistics, which requires a large number of data for optimization. Recently, prototypical network \cite{snell2017prototypical} shows potential on few-shot NER. The basic idea is to learn prototypes for each predefined entity class and an \textit{other} class, then classify examples based on which prototypes they are closest to \cite{fritzler2019few}. Most existing studies focus on the predefined classes and leverage the label semantic to reveal their dependency for enhancement \cite{hou2020few}. However, they ignore the massive semantics hidden in the words of other class (O-class for short).

\begin{figure}[t]
\centering
  \includegraphics[width=1\columnwidth]{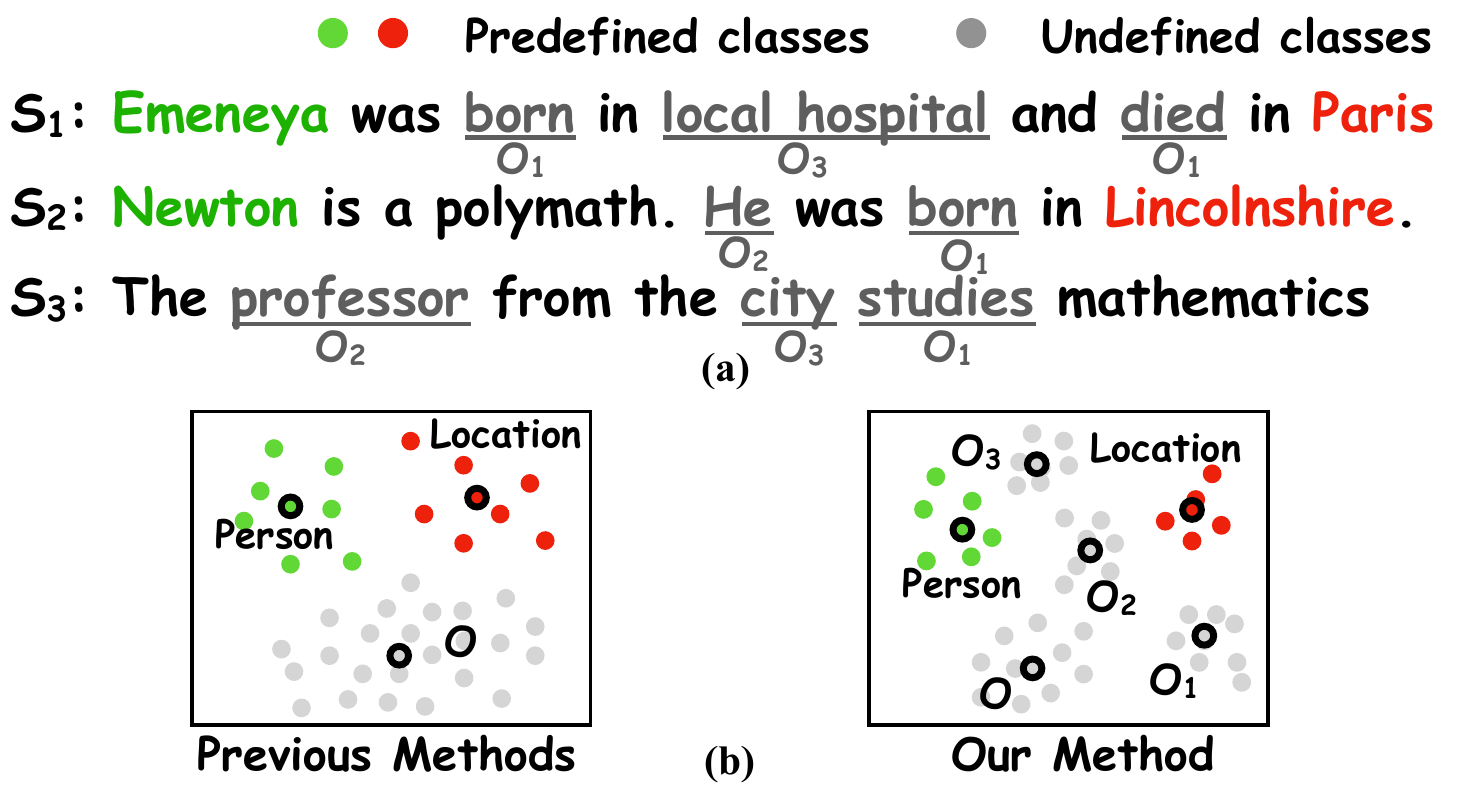}
  \caption{(a): Examples for undefined classes. (b): Different ways to handle O class (single prototype vs. multiple prototypes).}
  \label{fig:image-example}
\end{figure}

In this paper, we propose to learn from O-class words, rather than using predefined entity classes only, to improve few-shot NER. In fact, O-class contains rich semantics and can provide stand-by knowledge for named entity identification and disambiguation. As shown in Figure \ref{fig:image-example}(a), if we can detect an undefined class consisting of references to named entities (such as pronouns), then due to their interchangeability \cite {katz1963structure}, we will obtain prior knowledge for named entity identification. For example, \textit {Newton} can be replaced with \textit {he} or \textit {professor} in $S_2$ and $S_3$. If we can detect additional classes, including \textit {he} and \textit {professor}, we will have more evidence about where \textit{Newton} may appear. In addition, if we can detect an undefined class that composed of \textit{Action} ($O_1$), we may capture underlined relations between different named entities, which is important evidence when distinguishing the named entity type \cite{ghosh2016feature, zheng2017joint}. 


Nevertheless, it is challenging to detect related undefined classes from O class words due to two reasons: 1) Miscellaneous Semantics. O-class contains miscellaneous types of words. Based on our observations, although there are massive related yet undefined classes, the noise maybe even more, such as function and stop words. These noisy classes have little or negative impacts on the identification of target entities. Therefore, how to distinguish noise from task-related classes is a key point. 2) Lack of Golden Label. We neither have the labeled examples nor the metadata of each undefined class. The zero-shot methods \cite{DBLP:journals/corr/abs-1712-05972} fail in this case, since they need metadata (such as class name and class description) as known information. Unsupervised clustering methods also cannot meet quality requirements as shown in our experiment.  

To handle the issues, we propose the \textbf{M}ining \textbf{U}ndefined \textbf{C}lasses from \textbf{O}ther-class (MUCO) model to leverage the rich semantics to improve few-shot NER. Instead of a single prototype, we learn multiple prototypes to represent miscellaneous semantics of O-class. Figure \ref{fig:image-example}(b) shows the difference between our method and previous methods. To distinguish task-related undefined classes without annotations, we leverage weakly supervised signals from predefined classes and propose a zero-shot classification method called Zero-shot Miner. The main idea is inspired by transfer learning in prototypical network. Prototypical network can be quickly adapted to new class B when pre-training on related base class A. The underlined reason is that if two classes (A and B) are task-related, when we make examples in A class to cluster in the space, the examples in B class also tend to cluster in the space, even without explicit supervision on class B \cite{koch2015siamese}. Based on this phenomenon, we first perform prototype learning on predefined classes to cluster words in predefined classes, and then regard words in O-class that also tend to cluster as the undefined classes. Specifically, we train a binary classification to judge whether clustering occurs between any two of the words. After that, we label the found undefined classes back into sentences to jointly recognize predefined and undefined classes for knowledge transfer. Our contributions can be summarized as follows:
\begin{itemize}
\item We propose a novel approach MUCO to leverage rich semantics in O class to improve few-shot NER. To the best of our knowledge, this is the first work exploring O-class in this task. 
\item We propose a novel zero-shot classification method for undefined class detection. In the absence of labeled examples and metadata, our proposed zero-shot method creatively use the weakly supervised signal of the predefined classes to find undefined classes. 
\item We conduct extensive experiments on four benchmarks as compared with five state-of-the-art baselines. The results under both 1-shot and 5-shots settings demonstrate the effectiveness of MUCO. Further studies show that our method can also be conveniently adapted to other domains.
\end{itemize}


\section{Related Work}
Few-shot NER aims to recognize new categories with just a handful of examples \cite{feng2018improving, cao2019low}. Four groups of methods are adopted to handle the low-resource issue: knowledge enhanced, cross-lingual enhanced, cross-domain enhanced, and active learning. Knowledge-enhanced methods exploit ontology, knowledge bases or heuristics labeling \cite{fries2017swellshark,tsai2017improving,ma2016label} as side information to improve NER performance in limited data settings, which suffer from knowledge low-coverage issue. Cross-lingual \cite{feng2018improving,rahimi2019massively} and cross-domain enhanced methods \cite{wang2018label,zhou2019dual} respectively use labeled data from a counterpart language or a different domain as external supervised signals to avoid overfitting. When the language or domain discrepancy is large, these two methods will inevitably face the problem of performance degradation \cite{huang2017zero}. Active learning methods \cite{wei2019cost} explicitly expand corpus by selecting the most informative examples for manual annotation, which need extra human-laboring. Different from previous methods, we focus on mining the rich semantics in the O class to improve few-shot NER.


\subsection{Prototypical Network}
Prototypical network \cite{snell2017prototypical}, initially proposed for image classification, has been successfully applied to sentence-level classification tasks, such as text classification \cite{sun2019hierarchical} and relation extraction \cite{gao2019hybrid}. However, there is a dilemma to adapt prototypical network for token-level classification tasks such as NER. Prototypical network assumes that each class has uniform semantic and vectors belong to the same class should cluster in the space. However, in NER, data in O class contain multiple semantics and thus violate the uniform semantic hypothesis in prototypical network. To handle the issue, \citet{deng2020meta} first trains a binary classifier to distinguish O class from other predefined classes, and then adopt traditional prototypical network methods, which suffers from pipeline error propagation. \citet{fritzler2019few} does not calculate the prototype of O class from data, but directly sets a hyper-parameter $b_o$ as the fake distance similarity and optimize $b_o$ during training, which still regards O class as a whole. On the contrary, we are the first to divide O class into multiple undefined classes and explicitly learn multiple spatially-dispersed prototypes for O class. 

\section{Methodology}
\label{sec:met}

\begin{figure*}[htbp]
\centering
  \includegraphics[width=0.98\linewidth]{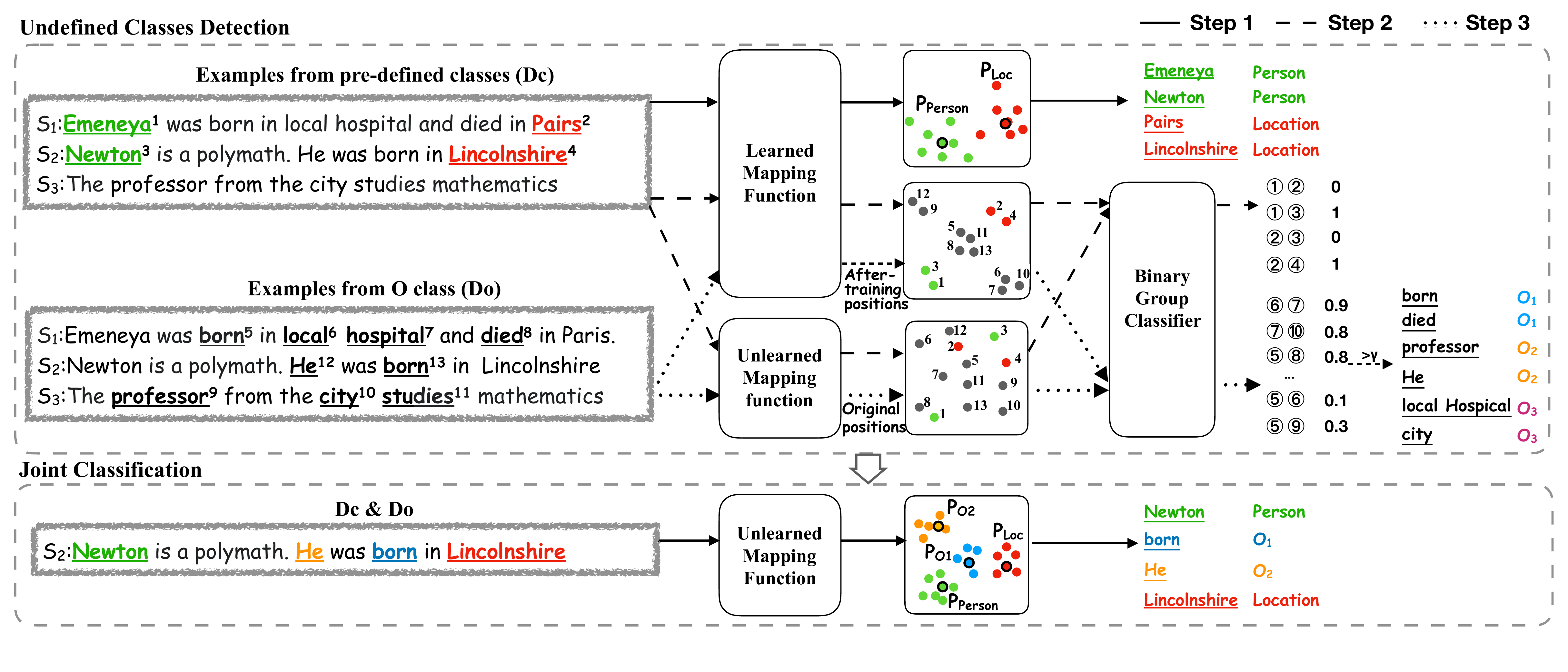}
  \caption{The architecture of the proposed MUCO model. We first detect undefined classes from O class, and then jointly classify the predefined classes and the found undefined classes for knowledge transfer. Specifically, in undefined classes detection, we propose a zero-shot classification method, which includes three steps. In step 1, we learn a mapping function through prototypical network training on predefined classes. In step 2, we learn a binary group classifier to judge whether any two points in predefined classes tend to cluster during the step 1 training. In step 3, we use the binary group classifier to infer pairs of examples in O class to distinguish multiple undefined classes.}
  \label{fig:architecture}
\end{figure*}

Figure \ref{fig:architecture} illustrated the architecture of the proposed MUCO model. MUCO is composed of two main modules: \textbf{Undefined Classes Detection} detects multiple undefined classes hidden in O class to fully exploit the rich semantics in O class. \textbf{Joint Classification} jointly classifies the undefined classes and predefined classes, so as to leverage the stand-by semantic knowledge in undefined classes to enhance the understanding of predefined classes.

\subsection{Notation}
In few-shot NER, we are given training examples $D=D_c\cup D_o$, where $D_c=\{x_i, y_i|_{i=1}^N\}$ is the training examples of predefined classes $C=\{c_1,c_2,\dots,c_k\}$ and $D_o=\{x_i|_{i=1}^M\}$ is the training examples of O class. For each example $(x,y)$, $x$ is composed by $S$ and $w_j$, where $S=<w_1,w_2,\dots,w_n>$ stands for the sentence and $w_{j}$ is the queried named entity, $y$ is the class label of the queried named entity $w_{j}$. We denote the prototype of class $y$ as $p_y$ and prototypes for all classes $C\cup O$ as $P=\{p_y|y\in C\cup O\}$. Formally, our goal is first to detect multiple undefined classes $O=\{o_1,o_2,\dots,o_r\}$ to label the examples in $D_o$, and then maximize the prediction probability $P(y|x)$ on $D_c$ and $D_o$.

\subsection{Undefined Classes Detection}
\label{ssc:undefined}
In few-shot NER, most of the words in the sentence belong to O class. Different from predefined classes, O class means none-of-the-above, and contains multiple undefined entity types. Previous methods ignore the fine-grained semantic information in O class and simply regard O as a normal class. We argue to further decouple O class into multiple undefined classes to fully exploit the rich semantics hidden in O class. 

In the section, we aim to detect undefined classes from O class. It is a non-trivial task since we lack metadata and golden labels to help us distinguish undefined classes. What is worse, the examples from O class is numerous and the search space is large. To handle the issue, we propose a zero-shot classification method called Zero-shot Miner to leverage the weak supervision from predefined classes for undefined classes detection. Our method inspires by transfer learning, we argue that if an undefined class is task-related, when we push the examples in predefined classes to cluster in the space, the examples in the undefined class should also have the signs of gathering, even without explicit supervision \cite{koch2015siamese}. For instance, in Figure \ref{fig:architecture}, if we guide \textit{Emeneya} and \textit{Newton} (the green points 1, 3) to cluster in the space, \textit{professor} and \textit{He} (the grey points 9, 12) will also tend to cluster in the space.

Based on this argument, undefined classes detection could be achieved by finding multiple groups of examples in O class that have a tendency to cluster during the training of the prototypical network on predefined classes. As shown in Figure \ref{fig:architecture}, there are three steps in our zero-shot classification method. In \textbf{step 1}, we train the prototypical network on predefined classes to obtain the learned mapping function. Through the learned mapping function the examples belonging to the same class will cluster in the space. In \textbf{step 2}, we train a binary group classifier on predefined classes base on the position features from the learned mapping function and unlearned mapping function to judge whether any two points tend to cluster during the step 1 training. In \textbf{step 3}, we use the learned binary group classifier in step 2 to infer examples in O class to distinguish undefined classes from each other. The following articles will illustrate the three steps sequentially.

 
\subsubsection{Step 1: Mapping Function Learning}
\label{ssec:pre}
In prototypical network, mapping function $f_\theta(x)$ aims to map the example $x$ to a hidden representation. BERT is adopted as the mapping function in our model, which is a pre-trained language representation model that employs multi-head attention as the basic unit, and have superior representation ability \cite{geng2019few}. 

We train the mapping function by correctly distinguishing the predefined classes. First, we extract the feature of the queried word. Formally, given the training example $(x,y) \in D_c$, where $x$ is composed of sentence $S=<w_1,w_2,\dots,w_n>$ and the queried word $w_j$, we extract the j-th representation of the sequence output of the last layer of BERT as the hidden representation.
\begin{equation}
   h = f_\theta(x)
\end{equation}


Then, following \cite{qi2018low}, we randomly initialize the prototype $p_y$ of class $y$ at the beginning of training, and then we shorten the distance between examples in class $y$ to prototype $p_y$ during training. Compared to traditional prototypical learning \cite{snell2017prototypical}, we do not need to waste part of the examples for prototype calculation.  
\begin{equation}
d(x,p_{y}) = -f_\theta(x)^Tp_{y}
\end{equation}
where $f_\theta(x)$ and $p_{y}$ are first normalized by L2 normalization. 

The final optimization goal for training the mapping function is
\begin{equation}
\label{eq:only_pre}
\begin{aligned}
    L(\theta_1) &=-log\frac{exp(-d(x,p_{y}))}{\sum_{p_c\in P_c}exp(-d(x,p_{c}))}
\end{aligned}
\end{equation}
where $P_c=\{p_{c} | c \in C\}$ stands for the prototypes of all the predefined classes.

\subsubsection{Step 2: Binary Group Classifier Training}
\label{ssb:find_multiple}
Recall that to detect multiple undefined classes, we need to find multiple example groups, and the examples in each group should have a tendency to cluster. 

To handle the issue, we learn a binary group classifier on predefined classes. The main idea is that if we can determine whether any two examples belong to the same group, we can distinguish groups from each other. Formally, given a pair of examples $(x_i,y_i)$ and $(x_j,y_j)$ in $D_c$, their original position $h_i$, $h_j$ from unlearned mapping function $f_\theta(x)$, and after-training position $\tilde{h}_i$, $\tilde{h}_j$ from learned mapping function $\tilde{f}_\theta(x)$, the probability of $x_i$ and $x_j$ belonging to the same class is defined as follows:
\begin{equation}
\label{equ_feature}
\begin{aligned}
b_{ij} = W([&h_i;h_j;\tilde{h}_i;\tilde{h}_j;|h_i-h_j|;\\&|\tilde{h}_i-\tilde{h}_j|;|h_i-\tilde{h}_i|;|h_j-\tilde{h}_j|])+b
\end{aligned}
\end{equation}
By comparing the distance variation between original positions $h$ and the after-training positions $\tilde{h}$, we can tell whether aggregation occurs between any of the two points. 


The optimization goal of the binary group classifier is
\begin{equation}
\begin{aligned}
L(\theta_2)= \frac{1}{N^2} \sum_i^N \sum_j^N &(-y_{ij}*log(b_{ij}) \\
&+(1-y_{ij})*log(1-b_{ij}))
\end{aligned}
\end{equation}
where $N$ is the numbers of the examples in predefined classes, and $y_{ij}$ is the label. If $x_i$ and $x_j$ are from the same predefined class ($y_i$=$y_j$), $y_{ij}$ is 1, otherwise 0. 

\subsubsection{Step 3: Binary Group Classifier Inference}
After training, we feed each pair of examples $x_u$ and $x_v$ in $D_o$ to the binary group classifier to obtain the group dividing results. The output $b_{uv}$ indicates the confidence that $x_u$ and $x_v$ belong to the same group. We set a threshold to divide the group. If $b_{uv}$ is larger than the threshold $\gamma$, $x_u$ and $x_v$ shall belong to the same group (undefined class). If consecutive words belong to the same group, we will treat these words as one multi-word entity. Noted that some of the examples in O class may not belong to any group. We assume that these examples come from the task-irrelevant classes, and no further classification is made for these examples. 


\para{Soft Labeling}
After the process of group dividing, we obtain labels of multiple undefined classes $O=\{o_1,o_2,\dots,o_r\}$. We further adopt the soft labeling mechanism. For each undefined class $o_i$, we calculate the mean of the examples as the class center, then we apply softmax on the cosine similarity between examples and its class center as the soft labels. Through soft labeling, we can consider how likely examples belong to the undefined classes.

\subsection{Joint Classification}
In the section, we take into consideration of both the predefined classes $C$ and the found undefined classes $O$ for joint classification. First, we label the examples in undefined classes back into the sentences, as shown in \textit{Joint Classification} of Figure \ref{fig:architecture}. Then, we optimize the examples to make them closer to the corresponding prototype for better discrimination. Comparing to the Equation \ref{eq:only_pre}, we add the prototypes from O class $P_o=\{p_{o_1},p_{o_2},\dots,p_{o_r}\}$ as candidate prototypes. 

Formally, given the examples $(x,y) \in D_c\cup D_o$, the corresponding prototype $p_y$ and prototypes set $P=P_c\cup P_o$ from both predefined classes $C$ and undefined classes $O$, the optimization object is defined as:
\begin{equation}
\begin{aligned}
\label{eq:whole}
    L(\theta_3) &= -log\frac{exp(-d(x, p_{y}))}{\sum_{p\in \{P_c\cup P_o\}}exp(-d(x,p))}
\end{aligned}
\end{equation}

\para{Scale Factor}
When calculating $d(x,p_y)$, the $f_\theta(x)$ and $p_{y}$ have been normalized and the value is limited to [-1, 1]. When softmax activation is applied, the output is unable to approach the one-hot encoding and therefore imposes a lower bound on the cross-entropy loss \cite{qi2018low}. For instance, even we give the golden prediction: giving 1 for correct category and -1 for the wrong ones, the probability of output $p(y|x) = e^1/[e^1 + (|C\cup T|-1)e^{-1}]$ is still unable to reach 1. The problem becomes more severe as we increase the number of named entity categories by introducing more categories for O class. To alleviate the issue, we modify Eq. \ref{eq:whole} by adding a trainable scalar $s$ shared across all classes to scale the inner product \cite{wang2017normface}.

\begin{equation}
\begin{aligned}
\label{eq:soft_whole}
    L(\theta_3) &= -log\frac{exp(-sd(x, p_{y}))}{\sum_{p\in \{P_c\cup P_t\}}exp(-sd(x, p))}
\end{aligned}
\end{equation}

\subsection{Implementation Details}
Following traditional prototypical network \cite{snell2017prototypical}, we pre-train the model on several base classes, whose types are disjoint to few-shot classes and have abundant labeled corpus. The underlined idea is to leverage existing fully annotated classes to improve the performance of the model on new classes with only a few annotations. All predefined classes (both base classes and few-shot classes) are used when searching for undefined classes, so that the annotations of undefined classes can be shared between pre-training and fine-tuning, which will improve the transfer performance of our model.

\section{Experiment}
\subsection{Datasets}
We conduct experiments on multiple datasets to reduce the dataset bias, including three English benchmarks Conll2003 \cite{sang2003introduction}, re3d \cite{dstl2017} and Ontonote5.0 \cite{pradhan2013towards} and one Chinese benchmark CLUENER2020 \cite{xu2020cluener2020}. Conll2003 contains 20,679 labeled sentences, distributed in 4 classes in the News domains. The data in re3d comes from defense and security domain, with 10 classes and 962 labeled sentences. Ontonotes5.0 has 17 classes with 159,615 labeled sentences in mixed domains - News, BN, BC, Web and Tele. CLUENER2020 has 10 fined grained entity types with 12,091 annotated sentences. For all of the datasets, we adopt BIO (\textbf{B}eginning, \textbf{I}nside, and \textbf{O}utside) labeling, which introduces an extra O class for non-entity words.
\begin{table*}[t]
\small
\centering
\caption{Overall Performance on Conll2003, re3d, Ontonote5.0 and ClUENER2020 dataset in 1-shot setting(\%).}
\label{tb:overall-1-shots}
\begin{tabular}{c|ccc|ccc|ccc|ccc}
\Xhline{2\arrayrulewidth}
\multirow{3}{*}{Methods} & \multicolumn{12}{c}{1-shot Named Entity Recognition}\\ \cline{2-13} & \multicolumn{3}{c}{Conll2003} & \multicolumn{3}{c}{re3d} & \multicolumn{3}{c}{Ontonote5.0} & \multicolumn{3}{c}{CLUENER2020} \\ \cline{2-13}
& P & R & F & P & R & F & P & R & F & P & R & F  \\
\Xhline{2\arrayrulewidth}
BERT & 61.00 & 50.46 & 54.28 & 31.49& 22.56& 26.13& 54.92 & 32.09 & 39.92 & 26.95 & 18.68 & 21.77\\
PN & 55.78 & 50.72 & 52.10 & 32.07 & 23.09 & 26.75 & 55.77 & 30.56 & 38.67 & 27.64 & 19.78 & 22.81\\
LTC & 78.19 & 70.36 & 73.31 & 29.84 & 19.33 & 23.34 & 60.83 & 43.25 & 50.04 & - & - & -\\
WPN & 77.87 & 86.58 & 81.40 & 43.12 & 38.90 & 40.27 & 58.29 & 54.39 & 56.20 & 76.63 & 70.96 & 73.50\\
MAML & 75.95 & 85.69 & 79.80 & 43.95 & 34.77 &37.83& 56.63 & 55.84 & 56.15 & 77.71 &69.53&73.08\\
\hline
MUCO (ours) & \textbf{81.70} & 83.98 &
\textbf{82.69} & 43.23 & \textbf{40.37} & \textbf{41.57} & 60.43 & 55.82 & \textbf{57.89} & \textbf{78.29} & \textbf{73.60} & \textbf{75.80} \\
\Xhline{2\arrayrulewidth}
\end{tabular}
\end{table*}

\begin{table*}[t]
\small
\centering
\caption{Overall Performance on Conll2003, re3d, Ontonote5.0 and ClUENER2020 dataset in 5-shot setting(\%).}
\label{tb:overall-5-shots}
\begin{tabular}{c|ccc|ccc|ccc|ccc}
\Xhline{2\arrayrulewidth}
\multirow{3}{*}{Methods} & \multicolumn{12}{c}{5-shots Named Entity Recognition}\\ \cline{2-13} & \multicolumn{3}{c}{Conll2003} & \multicolumn{3}{c}{re3d} & \multicolumn{3}{c}{Ontonote5.0} & \multicolumn{3}{c}{CLUENER2020} \\ \cline{2-13}
& P & R & F & P & R & F & P & R & F & P & R & F  \\
\Xhline{2\arrayrulewidth}
BERT &  73.94& 68.28 & 70.54 & 32.43  & 25.05 & 28.09 & 61.81 & 56.64 & 59.04 & 71.5& 68.14& 69.61\\
PN & 74.36 & 71.46 & 72.70 & 31.26 & 25.37 & 27.77 & 61.84 & 58.61 & 60.12  & 71.60 & 68.56 & 69.92\\
LTC & 85.89 & 82.41 & 83.97 & 40.98 & 32.00 & 35.83 & 62.06 & 46.08 & 52.35 &- & - & -\\
WPN & 94.39 & 95.00 & 94.68 & 40.93 & 38.63 & 39.68 & 65.28 & 67.66 & 66.34 & 80.52& 79.71& 80.04 \\
MAML & 94.76 & 96.04 & 95.37 &41.78 &39.49 &40.52 &65.99 &69.31 &67.57 &77.06&82.83&79.78\\
\hline
MUCO (ours) & \textbf{96.23} & 95.35 & \textbf{95.78} &  \textbf{43.04} & \textbf{41.70} & \textbf{42.37} & \textbf{73.27} & 69.00 & \textbf{71.06} & 78.88 & 82.67 & \textbf{80.64} \\
\Xhline{2\arrayrulewidth}
\end{tabular}
\end{table*}

\subsection{Data Split}
We divided the classes of each benchmark into two parts: base classes and few-shot classes. The few-shot classes for Conll / re3d / Ontonote / CLUENER are Person / Person, Nationality, Weapon / Person, Language, Money, Percent, Norp / Game, Government, Name, Scene. The rest are the base classes. The division is based on the average word similarity among classes (mean similarity is reported in Appendix A). At each time, the class with the largest semantic difference from other classes is selected and added to the few-shot classes until the number of few-shot classes reaches 1/3 of the base classes. In this way, we can prevent the few-shot classes and base classes from being too similar, leading to information leakage. We do not follow previous methods \cite{hou2020few} to adopt different datasets as base and few-shot classes, because there are overlapped classes in such data split, such as \textit{Person}, which will reduce the difficulty of few-shot setting. For base classes, all examples are used to train the base classifier. For few-shot classes, only K examples are used for training, and the rest are used for testing. Alternatively, we adopt the N-way K-shot setting for few-shot classes, where N is the number of few-shot classes and K is the number of examples sampled from each few-shot class. K is set to 1 and 5 respectively in our experiment. Noted that we can not guarantee the number of the examples is exactly equal to K when sampling, because there will be multiple class labels in one sentence. Following \cite{fritzler2019few}, we ensure there are at least K labels for each few-shot class. 

\subsection{Evaluation Metrics}
Following \cite{hou2020few}, we measure the precision, recall, and macro-averaged F1 scores on all few-shot classes. For fair comparison with baselines, as long as the found undefined class is classified as O class, it can be considered correct. We report the average on ten runs as the final results.

\subsection{Hyperparameters}
For feature extraction, we adopt BERT-base as our backbone \footnote{https://github.com/google-research/bert}, which has 12-head attention layers and 768 hidden embedding dimension. For learning rate, we adopt greedy search in the range of 1e-6 to 2e-4. We set learning rage to 2e-5 when pre-training on base classes and 5e-6 when fine-tuning on few-shot classes. The threshold $\gamma$ is set to 0.68 to ensure that the found undefined classes are sufficiently relevant to the predefined classes. The batch size is 128 and the maximum sequence length 128. We set the scale factor in Eq. \ref{eq:soft_whole} to 10 at the beginning. Our code is implemented by Tensorflow and all models can be fit into a single V100 GPU with 32G memory. The training procedure lasts for about a few hours. The best result appears around the 100 epochs of the training process. 

\subsection{Baselines}
We divide the baselines into two categories: 1) Supervised-Only Methods. \textbf{BERT} uses pre-trained BERT model to sequentially label words in sentence \cite{devlin2018bert}.  \textbf{Prototypical network (PN)} learns a metric space for each class \cite{snell2017prototypical}. Both of the methods are only trained on the few-shot classes.
2) Few-shot Methods. \textbf{L-TapNet+CDT (LTC)} uses semantic associations between base and few-shot classes to improve the prototype quality, which is only trained on base classes \cite{hou2020few}. We use the original published code \footnote{https://github.com/AtmaHou/FewShotTagging}. \textbf{Warm Prototypical Network (WPN)} \cite{fritzler2019few} is the transfer learning version of PN, which is first pre-trained on base classes and then fine-tuned on few-shot classes. \textbf{MAML} first learns fast-adapted parameters on base classes and then fine-tune the parameters on few-shot classes \cite{finn2017model}.
\subsection{Overall Performance}
\label{ssec:overall}
Table \ref{tb:overall-1-shots} and \ref{tb:overall-5-shots} present the overall performance of the proposed approach on four NER benchmarks - Conll2003, re3d, Ontonote5.0 and CLUENER2020. MUCO (ours) consistently outperforms state-of-the-art models, showing the effectiveness of exploiting the rich semantics in O class and the superiority of the proposed MUCO model. 

Compared with supervised-only methods (BERT and PN), few-shot methods (TransferBERT, WPN, MAML, L-TapNet+CDT and MUCO(ours)) achieve better performance. By first training on base classes, these methods will learn a prior, which prevents from overfitting densely labeled words. Among few-shot methods, our model achieves the best performance. Previous methods regard O class as a single class. On the contrary, we induce different undefined classes from O class, and add more task-related classes for joint training, which directly handles the dilemma of scarcity of data in few-shot learning and provides stand-by semantics to identify and disambiguate named entity, thereby improving the performance of few-shot NER. No matter English corpus (the first three) or Chinese corpus (the last one), our methods consistently improves the F score, showing the language-independent superiority of our method. Task-agnostic superiority also shows in section \ref{sst:task-agnostic ability}. Our undefined classes detection method is completely data-driven. The found undefined classes will be automatically adjusted to be useful and task-related based on current language or task predefined classes.

To further evaluate our core module \textit{undefined classes detection} in section \ref{ssc:undefined}, we introduce a \textit{Word-Similarity} (WS) baseline. WS detects undefined classes by performing KMeans \cite{kanungo2002efficient} in O words based on word similarity. To be fair, WS, like our method, uses soft-label enhancement (section \ref{ssb:find_multiple}). We report the final few-shot NER performance on Ontonote for comparison. 
\begin{figure}[ht]
\centering
  \includegraphics[width=0.75\linewidth]{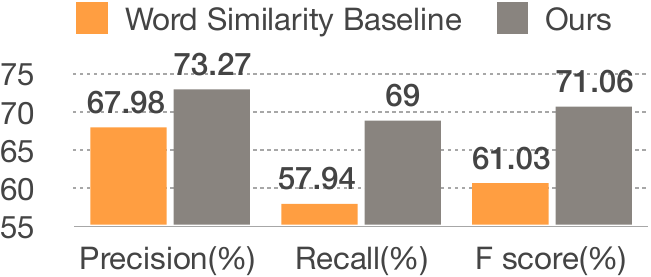}
  \caption{Few-shot NER Performance under Different Undefined Classes Detection Algorithm}
  \label{fig:ws}
\end{figure}

As shown in Figure \ref{fig:ws}, our method achieves better performance, which shows the superior of our undefined classes detection module. \textit {Word similarity} baseline only uses semantics of words and lacks weak supervision from predefined classes, so that noisy classes (such as punctuation) cannot be distinguished from task-related ones, which inevitably reduces the quality of undefined classes.
%

\subsection{Quality of Found Undefined Classes}
In the section, we evaluate the quality of the found undefined classes from quantitative and qualitative perspective. All the following experiments are conducted on Ontonote5.0.

For quantitative analysis, we invite three computer engineers to manually label 100 sentences for human evaluation. The metrics are Intra-class Correlation (IC) and Inter-class Distinction (ID). The IC statistics how many labels actually belong to the declared class. The ID counts how many labels belong to only one of the undefined classes, not to multiple classes. We obtain golden labels by applying the majority vote rule. Table \ref{tbl:HumanEvaluation} reports the average results on undefined classes.
\begin{table}[htpt]
  \small
  \centering
  \caption{Human Evaluation}
  \label{tbl:HumanEvaluation}
  \begin{tabular}{ccc}
    \Xhline{2\arrayrulewidth}
    Metrics & IC & ID \\
    \Xhline{2\arrayrulewidth}
    Average Score(\%) & 49.15 & 50.85\\
    \Xhline{2\arrayrulewidth}
  \end{tabular}
\end{table}

Considering the zero-shot setting, the accuracy of 49.15\% and 50.85\% is high enough, which indicates that the found undefined classes basically have semantic consistency within the classes and semantic difference between classes. 

\begin{table*}[ht]
  \small
  \centering
  \caption{Case Study of the Found undefined Classes}
  \label{tbl:examples of undefined classes}
  \begin{tabular}{c|l}
    \hline
     & Annotated Words \\
    \hline
    $O_1$ & gentleman; journalist; president; ambassador; I; he; they; businessmen from; and those Huwei people who;\\
    $O_2$ & the harbour; this land, which; over the river; with the great outdoors; outsides; to nature; the skyline; \\
    $O_3$ & some; a major; the small number; supplied; not only one of the; empty; large; increase of; was at the tail;\\
    $O_4$ & believe; comfort; attacked or threatened; arrest; geared; talks; not dealing; discussions; agreement;\\
    $O_5$ & stop; have; do; discussion; take; seek; sat down; negotiated; think; failed; replace;\\
    \hline
  \end{tabular}
\end{table*}

For qualitative analysis, we illustrate a case study in Table \ref{tbl:examples of undefined classes}. The words in $O_1$, $O_2$ and $O_3$ are mainly the general entity versions of Person, Location and Numerous respectively. According to the grammatical rules, general entities and named entities can be substituted for each other, \textit{Lincoln} can also be called \textit{president}, so identifying general entities can provide additional location knowledge and enhance named entity identification. The words in $O_4$ and $O_5$ are mainly \textit{Action}, which may imply relations between different named entities and provide important evidence for named entity disambiguation \cite{tong2020improving}. The errors mainly come from three aspects: 1) The surrounding words are incorrectly included, such as \textit{from} in \textit{businessmen from} in $O_1$; 2) Some strange words reduce intra-class consistency, such as \textit{was at the tail} in $O_3$; 3) There is semantic overlap between classes, such as $O_4 $ and $O_5 $. Future work will explore how to improve the quality of the undefined classes.


\subsection{Different Number of Undefined Classes}
Since our model needs to manually set the number of undefined classes, we observe the performance of the model under different number settings. We set the number of undefined classes to 1/2/5/10/25/50 by adjusting the threshold $\gamma$. 

\begin{figure}[htbp]
\centering
  \includegraphics[width=0.8\linewidth]{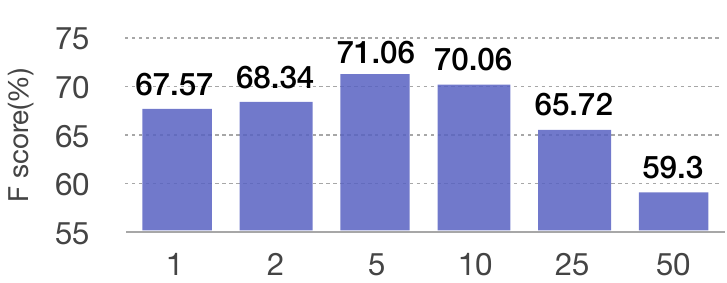}
  \caption{Different Numbers of Undefined Classes}
  \label{fig:number}
\end{figure}

Figure \ref{fig:number} illustrates the F score of MUCO (ours) on various numbers of undefined classes. It will impair the performance when the number is too large or too small. When the number is too large, the found classes will have overlapping problems, resulting in severe performance degradation (-11.51\%). When the number is too small, the model is unable to find enough task-related classes, limiting the ability to capture the fine-grained semantics in O class. Empirical experiments found that when the number of undefined classes is approximately equal to the number of few-shot classes, our method achieves the best performance (the number is 5 in Figure \ref{fig:number}). We argue that the number of predefined classes is proportional to the amount of information hidden in weak supervision. Therefore, with more predefined classes, we can also find more high-quality undefined classes.

\subsection{Cross-Domain Ability}
In this section, we answer whether our model could achieve superior performance facing the discrepancy of different domains. To simulate a domain adaption scenario, we choose the benchmark Conll2003 \cite{sang2003introduction} as the source domain and AnEM \cite{ohta2012open} as the target domain. The entity types in AnEM, such as \textit{Pathological-Formation}, are all medical academic terms and can ensure the discrepancy to common classes in Conll2003.

\begin{table}[H]
  \small
  \centering
  \caption{Domain Adaption Ability. }
  \label{tbl:cross-domain}
  \begin{tabular}{cccc}
    \Xhline{2\arrayrulewidth}
    Method & P& R& F \\
    \Xhline{2\arrayrulewidth}
    PN & 7.34 & 17.14 & 7.43 \\
    WPN & 33.06 & 31.95 & 26.90 \\
    \hline
    MUCO (Ours) & \textbf{34.17} & \textbf{32.84} & \textbf{28.31} \\
    \Xhline{2\arrayrulewidth}
  \end{tabular}
\end{table}

As illustrated in Table \ref{tbl:cross-domain}, our method achieves the best adaptation performance on the target domain. All the predefined classes, both in source domains and target domains, are used when detection undefined classes. The annotations of undefined classes can be shared between pre-training and fine-tuning, which will improve the transfer performance of our model. 



\subsection{Task-Agnostic Ability}
\label{sst:task-agnostic ability}
In this section, we answer whether our assumption of O class is task-agnostic and effective for few-shot token-level classification tasks other than NER. We conduct experiments on two tasks of widespread concern: Slot Tagging \cite{hou2020few} and Event Argument Extraction \cite{ahn2006stages}. Slot Tagging aims to discover user intent from task-oriented dialogue system. We adopt Snips dataset \cite{coucke2018snips} for Slot Tagging, and the split of train/test is We,Mu,Pl,Bo,Se/Re,Cr. Event Argument Extraction aims to extract the main elements of event from sentences. We adopt the ACE2005 dataset \footnote{http://projects.ldc.upenn.edu/ace/} with 33 classes and 6 domains. The train/test is bc,bn,cts,nw/un,wl.

\begin{table}[ht]
  \small
  \centering
  \caption{Task-Agnostic Effectiveness of Silent Majority(ours)}
  \label{tbl:task-agnostic}
  \begin{tabular}{cccc}
    \Xhline{2\arrayrulewidth}
    \multirow{2}{*}{Methods} & \multicolumn{3}{c}{ST} \\ \cline{2-4}
    & P & R & F \\
    \Xhline{2\arrayrulewidth}
    PN &  61.29 & 58.24 & 59.02 \\
    WPN &73.60 & 73.29 & 70.56 \\
    \hline
    Silent Majority (Ours) &  \textbf{75.92} & \textbf{73.71} & \textbf{72.04} \\
    \Xhline{2\arrayrulewidth}
  \end{tabular}
  \begin{tabular}{cccc}
    \Xhline{2\arrayrulewidth}
    \multirow{2}{*}{Methods} & \multicolumn{3}{c}{EAE} \\ \cline{2-4}
    & P & R & F\\
    \Xhline{2\arrayrulewidth}
    PN &  51.02 & 53.14 & 51.85\\
    WPN & 78.39 & 70.59 & 73.13\\
    \hline
    Silent Majority (Ours)  & \textbf{79.61} & \textbf{72.02} & \textbf{75.20}\\
    \Xhline{2\arrayrulewidth}
  \end{tabular}
\end{table}
    
As illustrated in Table \ref{tbl:task-agnostic}, the proposed model achieves superior performance on both tasks, which demonstrates the generalization ability of our method. No matter what task the predefined class belongs to, our method is always able to mine the task-related classes from the O class to help eliminate the ambiguity of the predefined class. The reason is that our detection method is entirely data-driven, and does not rely on manually writing undefined class descriptions. The found category will automatically change according to the task type of the entered predefined classes. Therefore, the migration cost between tasks of our method is meager.

\section{Conclusion}
In this paper, we propose \textbf{M}ining \textbf{U}ndefined \textbf{C}lasses from \textbf{O}ther-class (MUCO) to utilize the rich semantics in O class to improve few-shot NER. Specifically, we first leverage weakly supervised signals from predefined classes to detect undefined classes from O classes. Then, we perform joint classification to exploit the stand-by semantic knowledge in undefined classes to enhance the understanding of few-shot classes. Experiments show that our method outperforms five state-of-the-art baselines on four benchmarks. 

\section*{Acknowledgements}
This work is supported by the National Key Research and Development Program of China (2018YFB1005100 and 2018YFB1005101) and NSFC Key Project (U1736204). This work is supported by National Engineering Laboratory for Cyberlearning and Intelligent Technology, Beijing Key Lab of Networked Multimedia and the Institute for Guo Qiang, Tsinghua University (2019GQB0003). This research was conducted in collaboration with SenseTime. This work is partially supported by A*STAR through the Industry Alignment Fund - Industry Collaboration Projects Grant, by NTU (NTU–ACE2020-01) and Ministry of Education (RG96/20).

\bibliographystyle{acl_natbib}
\bibliography{acl2021}
\appendix
\section{Data Split}
We divided the classes of each benchmark into two parts: base classes and few-shot classes. The division is based on the average word similarity among classes. At each time, the class with the largest semantic difference from other classes is selected and added to the few-shot classes until the number of few-shot classes reaches 1/3 of the base classes. In this way, we can prevent the few-shot classes and base classes from being too similar, causing information leakage. The embedding of words are extracted from BERT, and the mean similarity is reported in Table \ref{word_similarity}.
\begin{table*}[h]
\caption{Mean Word Similarity between Predefined Classes}
\label{word_similarity}
\centering
\begin{tabular}{lllll}
\Xhline{2\arrayrulewidth}
\multicolumn{5}{c}{\textbf{Conll2003}}                                         \\
\Xhline{2\arrayrulewidth}
PER         & MISC       & LOC               & ORG          &                  \\
0.64        & 1.36       & 1.68              & 1.75         &                  \\
\Xhline{2\arrayrulewidth}
\multicolumn{5}{c}{\textbf{re3d}}                                              \\
\Xhline{2\arrayrulewidth}
Nationality & Person     & Weapon            & Temporal     & MilitaryPlatform \\
3.3         & 3.87       & 4.05              & 4.28         & 4.34             \\
Quantity    & Money      & DocumentReference & Location     & Organisation     \\
4.37        & 4.38       & 5.03              & 5.2          & 5.74             \\
\Xhline{2\arrayrulewidth}
\multicolumn{5}{c}{\textbf{Ontonotes5.0}}                                      \\
\Xhline{2\arrayrulewidth}
PERSON      & MONEY      & PERCENT           & LANGUAGE     & NORP             \\
0.31        & 2.2        & 3.88              & 4.14         & 4.49             \\
CARDINAL    & PRODUCT    & QUANTITY          & ORG          & LAW              \\
4.58        & 5.16       & 5.41              & 5.74         & 5.99             \\
TIME        & ORDINAL    & WORK\_OF\_ART     & GPE          & LOC              \\
6.01        & 6.23       & 6.24              & 6.7          & 7.18             \\
DATE        & FAC        & EVENT             &              &                  \\
7.27        & 7.53       & 8.11              &              &                  \\
\Xhline{2\arrayrulewidth}
\multicolumn{5}{c}{\textbf{ClUENER2020}}                                       \\
\Xhline{2\arrayrulewidth}
name        & government & game              & scene        & position         \\
7.28        & 7.38       & 7.43              & 7.45         & 7.6              \\
address     & movie      & company           & organization & book             \\
7.61        & 7.62       & 7.65              & 7.72         & 7.91\\
\Xhline{2\arrayrulewidth}
\end{tabular}
\end{table*}
 
\end{document}